\documentclass[10pt, a4paper]{article}
\usepackage{lrec2006}
\usepackage{graphicx}
\usepackage{url}

\title{Sockpuppet Detection in Wikipedia: A Corpus of Real-World Deceptive Writing for Linking Identities}

\name{Thamar Solorio, Ragib Hasan, Mainul Mizan}

\address{ University of Alabama at Birmingham \\
               Birmingham, Alabama\\
               solorio@cis.uab.edu, ragib@cis.uab.edu, mainul@cis.uab.edu\\}

\abstract{This paper describes the corpus of sockpuppet cases we gathered from Wikipedia. A sockpuppet is an online user account created with a fake identity for the purpose of covering abusive behavior and/or subverting the editing regulation process. We used a semi-automated method for crawling and curating a dataset of real sockpuppet investigation cases.  To the best of our knowledge, this is the first corpus available on real-world deceptive writing. We describe the process for crawling the data and some preliminary results that can be used as baseline for benchmarking research. The dataset will be released under a Creative Commons license from our project website: \url{http://docsig.cis.uab.edu}.  \\ \newline \Keywords{sockpuppet detection, authorship identification, deceptive language}}

\begin{document}

\maketitleabstract

\section{Introduction}
In Wikipedia, users can create multiple accounts for many
different purposes. According to Wikipedia's policies, each
user is supposed to create only one user account. However,
Wikipedia does not enforce the one-user-one-account rule
through technical means. As a result, users are free to create
multiple accounts if they want to. A secondary account created by a user for malicious purposes is called a sockpuppet. This ease of creating an identity has led malicious users to create multiple identities and use them for various purposes, ranging from block
evasion, false majority opinion claims, and vote stacking.

One of the main applications of the sockpuppet dataset is to develop an automated tool for sockpuppet detection in Wikipedia. Currently, the process for detecting sockpuppets is manual and involves significant experience from the administrators. In many cases, the user IP addresses have to be accessed by special Wikipedia administrators with IP-address viewing privileges (``checkusers''). This violates user privacy. Without accessing the IP addresses, the administrators need to depend on their experience in dealing with sockpuppets to detect similarities in writing style and behavior manually. That leaves a lot of room for error. In contrast, an automated tool trained using our sockpuppet dataset can be used to identify the sockpuppets without requiring IP address information or expert administrator knowledge. In practice, the automated tool can be used to assist administrators to more accurately identify malicious sockpuppets.

Besides the use in development of tools for automated detection of sockpuppets in Wikipedia, the sockpuppet dataset has many other potential applications. In particular, this corpus can be used by researchers working on authorship attribution problems. The sockpuppet corpus provides a real world data set of short messages from real malicious users. The sockpuppet cases involve text from actual users who are intentionally creating multiple identities and actively trying to hide their connections to the sockpuppet master. Therefore, using this corpus, researchers can test their work in a real life setting. This type of authorship attribution of short text has potential applications in identifying terrorists in web forums, online discussion boards, phone text messages, tweets and other social media interactions where comments and text tend to be brief and short in length.

\section{Related Work}
Authorship analysis has received a great deal of attention in recent years \cite{Stamatatos:08}. The field has grown from a pure manual stylistic analysis to machine learning approaches that combine stylistic features with richer representations of writing preferences, such as n-grams of syntactic features \cite{SidorovEtAl:13} and local histograms of character n-grams \cite{EscalanteEtAl:11a}. Recent work started exploring the limits of automated approaches to the problem of authorship analysis by looking at extremely short documents \cite{LaytonEtAl:10}, very large candidate sets \cite{KoppelEtAl:11}, and cross-domain scenarios \cite{Goldstein-StewartEtAtl:08}.

Less work has been devoted to authorship analysis on deceptive writing. Some of the exceptions include the work in \cite{BrennanEtAl:12,NovakEtAl:04}. The main barrier to study attribution in adversarial scenarios is the lack of suitable data. This is understandable as the nature of the problem makes it difficult to have readily available data where subjects have been intentionally trying to deceive humans. To solve this barrier researchers have turn to the generation of artificial data sets. For instance Novak et al. generated sub aliases from message boards by randomly splitting data from the same alias \cite{NovakEtAl:04}. Then they evaluated performance of their method on linking the two sub aliases. The Brennan-Greenstadt adversarial stylometry corpus was collected from volunteers \cite{BrennanEtAl:12}. The authors instructed the subjects to submit original writings of an academic nature. Then the subjects were asked to obfuscate their writing style during the creation of a topic specific writing of 500 words. In addition, subjects were also requested to submit an imitation writing excerpt, where they were instructed to imitate the writing of Cormac McCarthy in \textit{The Road}. Here again, the topic of the imitation writing was controlled by the corpus developers.

Both resources are valuable in that they enabled researchers to explore attribution approaches and allowed them to show that in adversarial scenarios state of the art approaches will degrade performance. This gap in performance calls for more research in deceptive writing. However, these two data sets still have an artificial flavor to them since the authors were not self motivated  and it is not clear whether this will cause major differences in the final stylistic markers of their writings. The sockpuppet corpus we created is a real-world alternative to the study of deceptive writing in social media. The authors were not aware of someone collecting their writings to study attribution, thus this new data set will allow the study of deceptive writing in the wild.

\section{Sockpuppet Investigations (SPI) in Wikipedia}
Wikipedia allows any editor to request investigation of suspected sockpuppetry. The requester needs to include any evidence of the abusive behavior. Typical evidence includes information about the editing patterns related to those accounts, such as the articles, the topics, vandalism patterns, timing of account creation, timing of edits, and voting pattern in disagreements. 

Once a case is filed, an administrator will investigate the case. An administrator is an editor with privileges to make account management decisions, such as banning an editor. The administrator performs a behavioral evidence investigation and will try to determine whether the two accounts are related and will then issue a decision confirming or rejecting the sockpuppetry case, or request involvement of a check user. Check users are higher privileged editors, who have access to private information regarding editors and edits, such as the IP address from which an editor has logged in. Check users perform a technical evidence investigation. But as explained in Wikipedia SPI description, these users will be involved in the investigation, if needed, only after strong behavioral evidence has been collected. 

When an SPI concludes with a confirmed sock puppetry verdict, the sockpuppet account will be banned indefinitely. The administrators have the discretion to establish bans or to block the main account as well.

The process to resolve SPI described above is time consuming and expensive. The last time we checked the list of current cases, on 10/23/13, there were 30$+$ unique SPI cases listed for the month of October. This high rate of cases filed in a single month show the need for a streamlined process to handle SPIs. The data set we created is a first step on this direction.

\section{Data Collection Process}
All the data we collected from Wikipedia is readily available from the Wikipedia website. Wikipedia archives all information related to each sockpuppet case filed, and once a verdict is issued, that too is stored in the archives.  However, because of the lack of a standard format in the archives, our process for data collection is semi-automated. The sockpuppet cases we collected were crawled from the following urls:
\begin{itemize}
\item \url{https://en.wikipedia.org/wiki/Wikipedia:Sockpuppet_investigations/SPI/Closed/2009}
\item \url{https://en.wikipedia.org/wiki/Wikipedia:Sockpuppet_investigations/SPI/Closed/2010}
\item \url{http://en.wikipedia.org/w/index.php?title=Wikipedia:Sockpuppet_investigations/Cases/Overview&offset=&limit=500&action=history}
\end{itemize}

For each case selected for addition to our corpus we collect all data from the talk pages of each editor involved in the SPI case. This step is done automatically by crawling the corresponding Wikipedia archives. We only collect data from discussion pages since these are free form discussions among editors that give editors more freedom to show their stylistic writing markers. In contrast, the basic namespaces in Wikipedia, and in particular the articles the editors contribute to, have a more restrictive format that can make difficult the identification of editors. Moreover, some of the edits in the main Wikipedia articles include things like reverts, or typo corrections, that are related to the user behavior and not necessarily to editors writing styles. Our main goal to develop this corpus is to support research in deceptive writing, and thus the behavior treats mentioned above fall outside this goal. However, this information could still be crawled at a later stage and be leveraged to perform a persona identification.

The manual process for this task involves retrieving the final decision reached by the investigative administrator or check user. There is no fixed format for recording decisions on SPI cases and therefore parsing the data with regular expressions will not work for most cases. We were required to visit each SPI case and read the discussion of any administrators investigating the case and check users involved. This was the bottle neck for the process and what prevented us from having a larger sample. Although we continue to add cases to our data set as feasible.

The majority of the SPI cases in Wikipedia end up being confirmed as sock puppets. This is reasonable since editors file cases after they have already seen some suspicious behavior. Therefore, to provide a larger number of non-sock puppet cases, we crawled pairs of editors that have not been involved in SPI before but that have participated in the same talk pages as editors involved in SPI cases.

\section{The Sockpuppet Corpus} 
 
 \begin{table*}[ht]
\centering
\small
\begin{tabular}{| p{11cm} |  }
\hline
\textbf{Comment from the sockpuppeteer: -Inanna-} \\ 
Mine was original and i have worked on it more than 4 hours\textbf{.I} have changed it many times by opinions\textbf{.L}ast one was accepted by all the users(except for khokhoi)\textbf{.I} have never used sockpuppets\textbf{.P}lease dont care Khokhoi,Tombseye and Latinus.They are changing all the articles about Turks.The most important and famous people are on my picture.\\ \hline
\textbf{Comment from the sockpuppet: Altau} \\ 
Hello\textbf{.I }am trying to correct uncited numbers in Battle of Sarikamis and Crimean War by resources but khoikhoi and tombseye always try to revert them\textbf{.C}ould you explain them there is no place for hatred and propagandas, please? \\ \hline
\textbf{Comment from another editor: Khoikhoi} \\ 
Actually, my version WAS the original image. Ask any other user. Inanna's image was uploaded later, and was snuck into the page by Inanna's sockpuppet before the page got protected. The image has been talked about, and people have rejected Inanna's image (see above).\\ \hline
\end{tabular}
\caption{Sample excerpt from a single sockpuppet case. We show in boldface some of the stylistic features shared between the sockpuppeter and the sockpuppet.}
\label{tab:sample}
\end{table*}
We originally collected around 700 cases, but after manual inspection we removed about 80 cases where editors did not have content on the talk pages. These were editors that just made contributions directly to Wikipedia pages but did not engage in any side discussions about them. The resulting corpus currently has 623 cases where 305 of them were confirmed SPI cases by Wikipedia administrators or check users. The remaining 318 are non-sockpuppet cases that combine 105 SPI cases where the administrators verdict was negative, and 213 cases we created from other editors.

Examples from a couple of cases are shown in Table~\ref{tab:sample}. In that table we show a comment from the editor named Inanna that was accused of being the puppeteer of editor Altau. For comparison purposes we show as well a comment made by another editor, not involved in the SPI case on the same talk pages. A noticeable feature in the table is the omission of a white space after the periods.

The table also shows that the comments resemble what we would see in web forum data. For our corpus we found out that the average length in characters is 529. While texts are short, previous work has carried out author identification from tweets \cite{LaytonEtAl:10}, and many researchers, ourselves included, have reached good prediction performance on social media data that is very similar to the data of this corpus. Some statistics about this dataset are shown in Table~\ref{t:stats}. 

\begin{table}
\centering
\footnotesize
\begin{tabular}{|  p{6cm} |r| }
\hline
Confirmed SPI cases & 305\\ \hline
Denied SPI cases & 105\\ \hline
Created non-sock puppet cases & 213\\ \hline
Average number of comments per case & $\sim$ 180 \\ \hline
Average number of comments per editors & $\sim$ 83 \\ \hline
\end{tabular}
\caption{The sockpuppet data set}
\label{t:stats}
\end{table}

\section{A Machine Learning Approach to Sockpuppet Detection}

\begin{figure}
\centering
\includegraphics[width=0.55\textwidth]{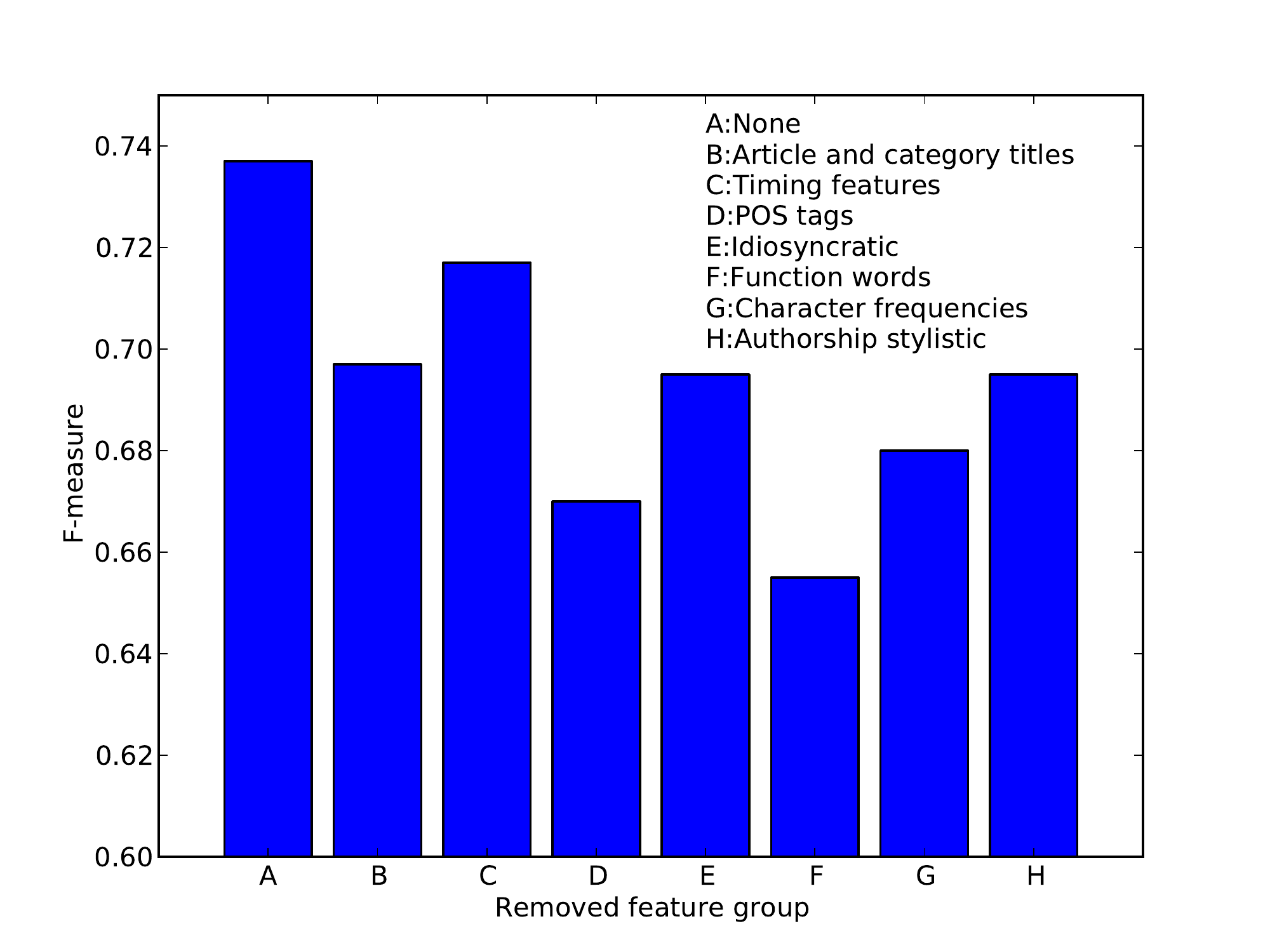}
\caption{The bars show average F-measures when testing support vector machine removing one feature group at a time in a 10 fold cross-validation setting.}
\label{f:exp}
\end{figure}

Earlier this year we did a case study of adapting a standard machine-learning authorship attribution approach to predict sockpuppet cases \cite{SolorioEtAl:13}. This preliminary study shows some promising results for this task. But it was based on a smaller set of cases, only 77. These 77 cases are a subset of the editors included in the new version of the corpus.

Here we present new results using all 623 cases in a ten-fold cross-validation setting. We hope these results can be used as a sort of baseline comparison for other researchers using this data set. 

For these experiments we also changed the underlying framework for the task. Here we assume any pair of editors can be considered an instance of the classification problem, a SPI, and the learner has to decide whether to declare the editors as belonging to the same person or not based on observations from the comments made by each editor involved. The features used in this problem are then the pairwise normalized differences of the feature vectors representing each comment. A complete list of features can be found at the following link: \url{https://www.dropbox.com/s/15tztqd48jrbr2h/features.list} and a detailed description is in our previous paper  \cite{SolorioEtAl:13}. Figure~\ref{f:exp} shows the results of training a support vector machine (SVM) classifier removing one feature group at a time. We used Weka's implementation of SVMs with default parameters. The best results (F-measure 73\%) are achieved using all features. These results are very similar to the results attained on our case study (F-measure 72\%).

\section{Conclusion}
This paper presents a new dataset that will enable research in authorship attribution under real-world adversarial conditions.  The nature of the data is very similar to what can be found in social media, which makes it an even more attractive resource as security and privacy concerns in social media data will continue to grow. The prediction results reported here will also be a good baseline for future research.

The data set will be available from the project website under a Creative Commons license. Our goal is to continue adding SPI cases on a regular basis to maintain an updated resource.
 
\bibliographystyle{lrec2006}
\bibliography{aa}

\end{document}